\documentclass{article}

\usepackage{arxiv}

\usepackage[utf8]{inputenc} 
\usepackage[T1]{fontenc}    
\usepackage{hyperref}       
\usepackage{url}            
\usepackage{booktabs}       
\usepackage{amsfonts}       
\usepackage{nicefrac}       
\usepackage{microtype}      
\usepackage{lipsum}

\usepackage{bm}
\usepackage{amsmath}
\usepackage{amssymb}
\usepackage{amsfonts}
\usepackage{graphicx}
\def\vec#1{\mathbf{#1}}
\usepackage{algorithm}
\usepackage{algorithmic}

\title{Neural Dynamic Mode Decomposition for End-to-End Modeling of Nonlinear Dynamics}

\author{Tomoharu Iwata\\ 
  NTT Communication Science Laboratories, Kyoto, Japan
  \And Yoshinobu Kawahara\\  
  Institute of Mathematics for Industry, Kyushu University, Fukuoka, Japan,\\
  Center for Advanced Intelligence Project, RIKEN, Tokyo, Japan}
\date{}

\begin{document}
\maketitle

\begin{abstract}
Koopman spectral analysis has attracted attention for understanding nonlinear dynamical systems by which we can analyze nonlinear dynamics with a linear regime by lifting observations using a nonlinear function. For analysis, we need to find an appropriate lift function. Although several methods have been proposed for estimating a lift function based on neural networks, the existing methods train neural networks without spectral analysis. In this paper, we propose neural dynamic mode decomposition, in which neural networks are trained such that the forecast error is minimized when the dynamics is modeled based on spectral decomposition in the lifted space. With our proposed method, the forecast error is backpropagated through the neural networks and the spectral decomposition, enabling end-to-end learning of Koopman spectral analysis. When information is available on the frequencies or the growth rates of the dynamics, the proposed method can exploit it as regularizers for training. We also propose an extension of our approach when observations are influenced by exogenous control time-series. Our experiments demonstrate the effectiveness of our proposed method in terms of eigenvalue estimation and forecast performance.
\end{abstract}

\section{Introduction}

The data-driven analysis of nonlinear dynamics
is important in a wide variety of fields,
such as sociology~\cite{guastello2013chaos}, epidemiology~\cite{liu1987dynamical},
neuroscience~\cite{bullmore2009complex}, physics~\cite{braun1998nonlinear}, finance~\cite{mann2016dynamic},
biology~\cite{daniels2015efficient}, and engineering~\cite{leonessa2001nonlinear}.
Recently, an approach based on
the Koopman operator~\cite{koopman1931hamiltonian,mezic2005spectral} has
garnered attention.
With this approach, a nonlinear dynamical system is lifted to the corresponding linear one
in a possibly infinite-dimensional space by a nonlinear function,
by which various methods that were developed for analyzing and controlling linear dynamical systems
can be used for nonlinear ones.
For example, such properties of nonlinear dynamical systems as frequencies and growth rates
can be identified with the eigenvalues of a Koopman operator in the lifted space.

One representative data-driven method based on the Koopman operator 
is dynamic mode decomposition
(DMD)~\cite{schmid2010dynamic,rowley2009spectral,kutz2016dynamic},
which analyzes the properties of nonlinear dynamics based on
the eigenvalues and eigenvectors of a low-rank approximation of a state transition matrix.
However, since DMD assumes that observations are obtained in a lifted space,
an appropriate lifted space must be defined.
To automatically find a lifted space,
several methods have been proposed that use neural networks
for modeling lift functions~\cite{takeishi2017learning,lusch2018deep,yeung2019learning,lee2020model,otto2019linearly}.
They train neural network-based autoencoders and linear dynamics in the lifted space,
where neither a low-rank approximation nor eigen decomposition is used.
After the autoencoders are trained, the dynamics is analyzed by the eigen decomposition
of the estimated state transition matrix as post processing.
Therefore, eigenvalues and eigenvectors, which contain important information for characterizing the dynamics,
are not directly used for training the autoencoders.

In this paper, we propose neural dynamic mode decomposition (NDMD), which trains
autoencoders such that the forecast error is minimized
when the dynamics is analyzed by DMD in lifted space.
With NDMD, a forecast is made based on the following procedure.
First, observations are encoded by a neural network-based encoder in a lifted space,
where the encoder approximates a lift function.
Second, a forecast is made in the lifted space by applying DMD to the encoded data.
Finally, the forecasted encoded data are decoded by a neural network-based decoder in the observation space.
With DMD, a low-rank approximation based on singular value decomposition (SVD) and eigen decomposition are used for the forecast.
The procedure, including SVD and eigen decomposition, resembles a single neural network model.
Since SVD~\cite{giles2008collected} and eigen decomposition~\cite{van2007computation,boeddeker2017computation}
are differentiable, we can backpropagate the forecast error through the model
and learn a model that includes DMD analysis in an end-to-end fashion.
The model is trained by minimizing the forecast error by a stochastic gradient descent method.

Since NDMD estimates the eigenvalues in the middle of the layers,
when the frequencies and/or the growth rates of the dynamics are known,
we can use the auxiliary information for training the model as regularizers.
We can also extend NDMD for analyzing dynamics when exogenous control time-series data are given,
where observation time-series data are affected by the exogenous time-series data.

The following are the main contributions of this paper:
\begin{enumerate}
    \item We propose a neural network-based method that trains a nonlinear lift function for Koopman spectral analysis such that DMD in a lifted space properly models the dynamics of the given time-series data.
    \item We present an extension of the proposed method to improve the performance using auxiliary information on the frequencies and/or the growth rates.
    \item We present an extension of the proposed method for training with exogenous control time-series data.
    \item We experimentally confirm that the proposed method can appropriately estimate eigenvalues and forecast future time-series better than the existing methods.
\end{enumerate}
The remainder of this paper is organized as follows.
In Section~\ref{sec:related},
we briefly describe related work.
In Section~\ref{sec:preliminary},
we explain Koopman spectral analysis,
on which our proposed method is based.
In Section~\ref{sec:proposed}, we propose NDMD for the end-to-end learning of DMD analysis in a lifted space.
We also propose NDMD when auxiliary information is available
and when exogenous control time-series are given.
In Section~\ref{sec:experiments}, we demonstrate the effectiveness of the proposed method
in terms of the eigenvalue estimation and forecasting performance.
Finally, we present concluding remarks and discuss future work in Section~\ref{sec:conclusion}.

\section{Related work}
\label{sec:related}

With DMD, we need to manually prepare an appropriate lift function according to the underlying nonlinear dynamics.
For modeling lift functions, a number of methods have been proposed,
such as basis functions~\cite{williams2015data,dietrich2020koopman}
and kernels~\cite{williams2015kernel,kawahara2016dynamic}.
These methods work well only if appropriate basis functions or kernels are prepared.
For learning lift functions from data,
several neural network-based methods have been
proposed~\cite{takeishi2017learning,yeung2019learning,lee2020model,otto2019linearly,xie2020linearization}.
However, these methods do not perform eigen decomposition during training.
On the other hand, the proposed method
learns a lift function such that the forecasting performance is improved
when the eigenvalues and the eigenvectors in the lifted space are used for forecasting.
\cite{lusch2018deep} used an auxiliary neural network for modeling eigenvalues.
On the other hand, the proposed method directly calculates eigenvalues without auxiliary neural networks.
Although DMD with exogenous control time-series has been proposed~\cite{proctor2016dynamic},
it does not learn nonlinear lift functions.

\section{Preliminaries: Koopman spectral analysis}
\label{sec:preliminary}

This section explains Koopman spectral analysis, which is the underlying theory of the proposed method.
We consider the following discrete-time nonlinear dynamical system in (latent) state space $\mathcal{W}$:
\begin{align}
\vec{w}_{t+1}=q(\vec{w}_{t}), \quad \bm{\psi}_{t}=h(\vec{w}_{t}),
\end{align}
where $\vec{w}_{t}\in\mathcal{W}$ is the state vector at timestep $t\in\mathbb{N}$,
$q:\mathcal{W}\rightarrow\mathcal{W}$ is a nonlinear state transition function,
$\bm{\psi}_{t}\in\mathbb{R}^{K}$ is the lifted observation vector at timestep $t$, and
$h:\mathcal{W}\rightarrow\mathbb{R}^{K}$ is a nonlinear observation function.
The Koopman operator~\cite{koopman1931hamiltonian}, denoted by $\mathcal{K}$, is an infinite-dimensional linear operator, defined for any $\vec{w}_t\in\mathcal{W}$ by
\begin{align}
\mathcal{K} h(\vec{w}_{t})=h(q(\vec{w}_{t})).
\end{align}

\section{Proposed method}
\label{sec:proposed}

In Section~\ref{sec:ndmd}, we propose neural dynamic mode decomposition (NDMD),
which is a neural network-based model for the end-to-end learning of a low-dimensional approximation of the lift function
in Koopman spectral analysis.
In Section~\ref{sec:train}, we present NDMD's training procedure.
In Section~\ref{sec:ndmda}, we propose an extension of NDMD when the auxiliary information on the eigenvalues of dynamics is available.
In Section~\ref{sec:ndmdc}, we propose another extension of NDMD for learning a lift function
when observations are influenced by exogenous control time-series. 

\subsection{Neural dynamic mode decomposition}
\label{sec:ndmd}

Let $\vec{x}_{t}\in\mathbb{R}^{M}$ be an observation vector at timestep $t$, 
which is obtained by a set of observables
$\tilde{h}:=[h_1,\ldots,h_M]$ as $\vec{x}_t=\tilde{h}(\vec{w}_t)$. Since the space spanned by $\tilde{h}$ is not necessarily large as mentioned in the previous section, we consider a $K$-dimensional lifted space,
where observation vectors are mapped onto the lifted space
by a nonlinear function,
\begin{align}
  \bm{\psi}_{t}=f(\vec{x}_{t}),
  \label{eq:encoder}
\end{align}   
where $\bm{\psi}_{t}\in\mathbb{R}^{K}$ is the encoded vector at timestep $t$,
and $f$ is modeled by a neural network called an encoder.
That is, we expect that the space spanned by composition $f\circ \tilde{h}$ is sufficiently large (in the sense of \cite{tu2013dynamic}) for the underlying dynamics behind the data.
Thus, in the lifted space,
encoded vectors are assumed to follow a discrete linear dynamics:
\begin{align}
\bm{\psi}_{t+1}\approx\vec{A}\bm{\psi}_{t},
\end{align}
where $\vec{A}\in\mathbb{R}^{K\times K}$ is a linear transition matrix.
With the proposed method,
infinite-dimensional Koopman operator $\mathcal{K}$
is approximated by low-dimensional linear transition matrix $\vec{A}$.

Our model obtains the linear dynamics by applying DMD on the encoded vectors.
We define two matrices, $\bm{\Psi}_{1}$ and 
$\bm{\Psi}_{2}$, which are constructed from the encoded vectors: 
\begin{align}
  \bm{\Psi}_{1}=[\bm{\psi}_{\tau_{1}},\bm{\psi}_{\tau_{2}},
    \cdots,\bm{\psi}_{\tau_{S}}]\in\mathbb{R}^{K\times S},
  \quad
  \bm{\Psi}_{2}=[\bm{\psi}_{\tau_{1}+1},\bm{\psi}_{\tau_{2}+1},
  \cdots,\bm{\psi}_{\tau_{S}+1}]\in\mathbb{R}^{K\times S},
  \label{eq:Psi}
\end{align}
where $\bm{\Psi}_{1}$ consists of the encoded vectors at timesteps $\bm{\tau}=\{\tau_{1},\cdots,\tau_{S}\}$,
and $\bm{\Psi}_{2}$ consists of the encoded vectors at their next timesteps.
Timesteps $\bm{\tau}$ can be non-consecutive, and unordered.

The transition matrix can be estimated by the least squares method:
\begin{align}
\hat{\vec{A}}=\arg\min_{\vec{A}}\parallel\bm{\Psi}_{2}-\vec{A}\bm{\Psi}_{1}\parallel^{2}
=\bm{\Psi}_{2}\bm{\Psi}_{1}^{\dagger},
\label{eq:Ahat}
\end{align}
where $\dagger$ denotes the pseudo-inverse.
Our model considers
the low-rank approximation of the transition matrix, $\tilde{\vec{A}}=\vec{U}^{\top}\vec{A}\vec{U}$,
as it did with DMD.
Here, $\vec{U}\in\mathbb{R}^{N\times R}$ is obtained
by the singular value decomposition (SVD) of $\bm{\Psi}_{1}$:
\begin{align}
  \bm{\Psi}_{1}\approx\vec{U}\bm{\Sigma}\vec{V}^{\top},
  \label{eq:svd}
\end{align}
where $\top$ denotes the transpose,
$\vec{\Sigma}\in\mathbb{R}^{R\times R}$,
$\vec{V}\in\mathbb{R}^{S\times R}$,
and $R$ is the reduced rank.
Then an estimate of the low-ranked transition matrix is given:
\begin{align}
\tilde{\vec{A}}=\vec{U}^{\top}\bm{\Psi}_{2}\vec{V}\bm{\Sigma}^{-1}\in\mathbb{R}^{R\times R}.
\label{eq:Atilde}
\end{align}
Using the low-rank approximation,
we can reduce the noise in the encoded vectors for modeling the dynamics.

Let the columns of $\vec{Y}\in\mathbb{C}^{R\times R}$ be the eigenvectors of $\tilde{\vec{A}}$, where
$\bm{\Lambda}\in\mathbb{C}^{R\times R}$ is a diagonal matrix containing corresponding eigenvalues $\lambda_{k}$:
\begin{align}
\tilde{\vec{A}}\vec{Y}=\vec{Y}\bm{\Lambda}.
\label{eq:eigen}
\end{align}
The encoded vector at timestep $t$
is forecasted using the eigenvalues and the eigenvectors:
\begin{align}
  \hat{\bm{\psi}}_{t}=\bm{\Phi}\bm{\Lambda}^{t-\tau_{1}}\bm{\alpha}=\sum_{r=1}^{R}\lambda_{r}^{t-\tau_{1}}\alpha_{r}\bm{\phi}_{r},
  \label{eq:psihat}  
\end{align}
where
\begin{align}
  \bm{\Phi}=\bm{\Psi}_{2}\vec{V}\bm{\Sigma}^{-1}\vec{Y}\in\mathbb{C}^{K\times R},
\end{align}
is the DMD modes, and $\bm{\alpha}=\bm{\Phi}^{\dagger}\bm{\psi}_{\tau_{1}}\in\mathbb{C}^{R}$.
Using eigen decomposition,
we can efficiently obtain the estimation at any timestep 
since the power of diagonal matrix $\bm{\Lambda}^{t-1}$
is calculated by the power of its diagonal elements.
In addition, when auxiliary information on the eigenvalues and/or eigenvectors is available, we can use it as a regularizer
for training our model as described in Section~\ref{sec:ndmda}.
In Eq.~(\ref{eq:psihat}), the encoded vectors are decomposed into $R$ different dynamics,
where the dynamics of the $r$th component is defined by eigenvalue $\lambda_{r}$.

The observation vector at timestep $t$
is forecasted by mapping forecasted encoded vector $\hat{\bm{\psi}}_{t}$ into the observation space:
\begin{align}
  \hat{\vec{x}}_{t}=g(\hat{\bm{\psi}}_{t}),
  \label{eq:decoder}
\end{align}
where $g$ is the decoder modeled by a neural network.

\begin{figure*}[t!]
\centering
  \includegraphics[width=46em]{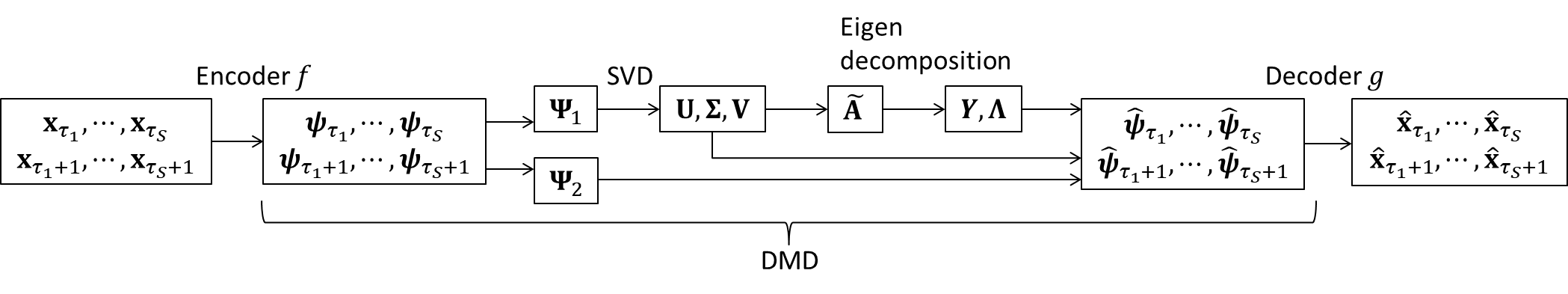}
\caption{Our model: First, observation vectors $\{\vec{x}_{t}\}$ are encoded to $\{\bm{\psi}_{t}\}$ in the lifted space by encoder $f$. Second, encoded vectors are forecasted based on DMD using SVD and eigen decomposition. Third, forecasted encoded vectors are decoded in the observation space by decoder $g$.}
\label{fig:model}
\end{figure*}

Our model including the SVD and eigen decomposition can be seen as a single neural network
that takes $\{\vec{x}_{t}\}_{t\in\vec{t}}$ as input,
and outputs forecasts $\{\hat{\vec{x}}_{t}\}_{t\in\vec{t}}$,
where $\vec{t}=\{\tau_{1},\tau_{2},\cdots,\tau_{S},\tau_{1}+1,\tau_{2}+1,\cdots,\tau_{S}+1\}$.
Figure~\ref{fig:model} shows the architecture of our model.
Since SVD~\cite{giles2008collected} and eigen decomposition~\cite{van2007computation,boeddeker2017computation}
are differentiable,
we can backpropagate the loss through our model for training.

The number of eigenvalues estimated by our model is at most $K$, which is the dimensionality of the lifted space.
We can easily limit the number of eigenvalues for obtaining sparse but essential dynamics
by tuning the number of output units of the encoder, as well as the number of input units of the decoder,
even if the dimensionality of the observation space is high.

\subsection{Training}
\label{sec:train}

Suppose that we are given time-series data of length $T$,
$[\vec{x}_{1},\cdots,\vec{x}_{T}]$.
The parameters in the encoder and decoder neural networks, $\bm{\Theta}$, are estimated
by minimizing the expected forecast error of our model in the observation space:
\begin{align}
\hat{\bm{\Theta}}=\arg\min_{\bm{\Theta}} L, \quad L=\mathbb{E}_{t}[\parallel \vec{x}_{t}-\hat{\vec{x}}_{t}\parallel^{2}],
\label{eq:Thetahat}
\end{align}
where $\mathbb{E}$ denotes the expectation.
In particular, we minimize the objective function using a stochastic gradient descent based method with batch size $S$,
as shown in Algorithm~\ref{alg:train}.

\begin{algorithm}[t!]
  \caption{Training procedures of the proposed method.}
  \label{alg:train}
  \begin{algorithmic}[1]
    \renewcommand{\algorithmicrequire}{\textbf{Input:}}
    \renewcommand{\algorithmicensure}{\textbf{Output:}}
    \REQUIRE{Time-series data $\vec{x}_{1},\cdots,\vec{x}_{T}$, batch size $S$}
    \ENSURE{Estimated neural network parameters $\hat{\bm{\Theta}}$}
    \STATE Initialize parameters $\bm{\Theta}$.
    \WHILE{End condition is satisfied}
    \STATE Randomly sample $S$ timesteps $\bm{\tau}=\{\tau_{1},\cdots,\tau_{S}\}$ from $\{1,\cdots,T-1\}$.
    \STATE Encode observation vectors $\bm{\psi}_{t}$ by Eq.~(\ref{eq:encoder}) for $t\in\vec{t}\equiv\bm{\tau}\cup\bm{\tau}+1$.
    \STATE Construct two encoded matrices, $\bm{\Psi}_{1}$ and $\bm{\Psi}_{2}$, by Eq.~(\ref{eq:Psi}).
    \STATE Take the SVD of $\bm{\Psi}_{1}$ in Eq.~(\ref{eq:svd}).
    \STATE Calculate $\tilde{\vec{A}}$ by Eq.~(\ref{eq:Atilde}).
    \STATE Take the eigen decomposition of $\tilde{\vec{A}}$ in Eq.~(\ref{eq:eigen}).
    \STATE Forecast encoded vectors $\hat{\bm{\psi}}_{t}$ by Eq.~(\ref{eq:psihat}) for $t\in\vec{t}$.
    \STATE Decode the forecasted encoded vectors to obtain forecasted observation vectors $\hat{\vec{x}}_{t}$ by Eq.~(\ref{eq:decoder}) for $t\in\vec{t}$.
        \STATE Calculate loss $L=\frac{1}{|\vec{t}|}
        \sum_{t\in\vec{t}}\parallel \bm{x}_{t}-\hat{\bm{x}}_{t}\parallel^{2}$,
        and its gradient.
    \STATE Update parameters $\bm{\Theta}$ using the loss and the gradient.
    \ENDWHILE
  \end{algorithmic}
\end{algorithm}

\subsection{Neural DMD with auxiliary information}
\label{sec:ndmda}

The real part of the eigenvalues
represents the growth and decay of the dynamics,
which diverges when $\mathrm{real}(\lambda_{k})>0$
and converges when $\mathrm{real}(\lambda_{k})<0$.
The imaginary part of the eigenvalues represents the frequency of the dynamics, $\frac{|\mathrm{imag}(\lambda_{k})|}{2\pi}$.
When auxiliary information on the dynamics is available,
we can use it for training our model
by adding regularizer $\eta$ in the objective function in Eq.~(\ref{eq:Thetahat}):
\begin{align}
L=\mathbb{E}_{t}[\parallel \vec{x}_{t}-\hat{\vec{x}}_{t}\parallel^{2}]+\beta\eta(\vec{E}),
\end{align}
where $\beta>0$ is a hyperparameter, $\eta$ is a regularization function,
and $\vec{E}$ is auxiliary information.
For example, suppose we are given true eigenvalues $\vec{E}=\bm{\lambda}^{*}=\{\lambda_{r}^{*}\}_{r=1}^{R}$
as auxiliary information.
The regularizer can be modeled such that estimated eigenvalues $\bm{\Lambda}$ in Eq.~(\ref{eq:eigen}) in our model become similar to the true ones:
\begin{align}
\eta(\bm{\lambda}^{*})&=\sum_{r=1}^{R}\min_{r'\in\{1,\cdots,R\}}|\lambda_{r}^{*}-\lambda_{r'}|
+\sum_{r'=1}^{R}\min_{r\in\{1,\cdots,R\}}|\lambda_{r}^{*}-\lambda_{r'}|,
\end{align}
where the first term is the absolute distance between true eigenvalue $\lambda^{*}_{r}$
and estimation $\lambda_{r'}$, which is closest to the true one,
and the second term is the absolute distance between estimated eigenvalue $\lambda_{r}$
and true $\lambda_{r'}$, which is closest to the estimated one.
We can incorporate other types of auxiliary information as regularizers.
For example, when only frequency information is available, a regularizer on the imaginary parts of the eigenvalues can be used.
When there is a limit cycle dynamics, we can use a regularizer that turns the real part of an eigenvalue to zero.

\subsection{Neural DMD with control}
\label{sec:ndmdc}

We propose an extension of neural DMD for analyzing dynamics using time-series data with control,
where we observe the exogenous variables that affect the dynamics.

Let $\vec{z}_{t}\in\mathbb{R}^{D}$ be the observed exogenous control vector at timestep $t$.
We encode $\vec{z}_{t}$ by nonlinear function $c$:
\begin{align}
\bm{\xi}_{t}=c(\vec{z}_{t}),
\end{align}
where $\bm{\xi}_{t}\in\mathbb{R}^{N}$ is the encoded exogenous vector at timestep $t$,
and $c$ is modeled by a neural network.
Our model assumes the following linear dynamics in the lifted space:
\begin{align}
\bm{\psi}_{t+1}\approx\vec{A}\bm{\psi}_{t}+\vec{B}\bm{\xi}_{t},
\label{eq:dynamics_c}
\end{align}
where the encoded observation vector at the next timestep is determined
by the encoded observation vector at the previous timestep and the encoded exogenous vector.

Our model obtains linear dynamics in the lifted space based on DMD with control~\cite{proctor2016dynamic}.
We define the following three matrices, $\bm{\Psi}_{1}$, 
$\bm{\Psi}_{2}$, and $\bm{\Xi}_{1}$,
from the encoded observation and exogenous vectors:
\begin{align}
  \bm{\Psi}_{1}&=[\bm{\psi}_{1},\bm{\psi}_{2},\cdots,\bm{\psi}_{T-1}]\in\mathbb{R}^{K\times (T-1)},\quad
  \bm{\Psi}_{2}=[\bm{\psi}_{2},\bm{\psi}_{3},\cdots,\bm{\psi}_{T}]\in\mathbb{R}^{K\times (T-1)},
  \nonumber\\
  \bm{\Xi}_{1}&=[\bm{\xi}_{1},\bm{\xi}_{2},\cdots,\bm{\xi}_{T-1}]\in\mathbb{R}^{N\times (T-1)}.
  \label{eq:Psi_c}
\end{align}
Note that $\bm{\Psi}_{1}, \bm{\Psi}_{2}$, and $\bm{\Xi}_{1}$ should be consecutive,
unlike Eq.~(\ref{eq:Psi}).
Let $\bm{\Omega}$ be a concatenation of $\bm{\Psi}_{1}$ and $\bm{\Xi}_{1}$:
\begin{align}
  \bm{\Omega}=[\bm{\Psi}_{1},\bm{\Xi}_{1}]^{\top}\in\mathbb{R}^{(K+N)\times(T-1)}.
  \label{eq:Omega}
\end{align}   
We take the SVD of $\bm{\Omega}$:
\begin{align}
  \bm{\Omega}\approx\vec{U}\bm{\Sigma}\bm{V}^{\top},
  \label{eq:svd_Omega}
\end{align}
where $\vec{U}\in\mathbb{R}^{(K+N)\times P}$,$\bm{\Sigma}\in\mathbb{R}^{P\times P}$,$\vec{V}\in\mathbb{R}^{P\times(T-1)}$,
and $P$ is the reduced rank.
We obtain estimates of $\vec{A}$ and $\vec{B}$
by least squares using the SVD result:
\begin{align}
  \hat{\vec{A}}=\bm{\Psi}_{2}\vec{V}\bm{\Sigma}^{-1}\vec{U}_{1}^{\top},
  \quad
  \hat{\vec{B}}=\bm{\Psi}_{2}\vec{V}\bm{\Sigma}^{-1}\vec{U}_{2}^{\top},
\end{align}
where $\vec{U}_{1}\in\mathbb{R}^{K\times P}$, $\vec{U}_{2}\in\mathbb{R}^{N\times P}$, and
$\vec{U}=[\vec{U}_{1},\vec{U}_{2}]^{\top}$.

Let $\hat{\vec{U}}\in\mathbb{R}^{K\times R}$, $\hat{\bm{\Sigma}}\in\mathbb{R}^{R\times R}$,
and $\hat{\bm{V}}\in\mathbb{R}^{(T-1)\times R}$
be the SVD of $ \bm{\Psi}_{2}$ with reduced rank $R$:
\begin{align}
  \bm{\Psi}_{2}\approx\hat{\vec{U}}\hat{\bm{\Sigma}}\hat{\bm{V}}^{\top}.
  \label{eq:svd_psi2}
\end{align}
Then the low-rank approximation of $\vec{A}$
is given:
\begin{align}
  \tilde{\vec{A}}=\hat{\vec{U}}^{\top}\bm{\Psi}_{2}\vec{V}\bm{\Sigma}^{-1}\vec{U}_{1}^{\top}\hat{\vec{U}}.
  \label{eq:Atilde_c}
\end{align}
The dynamics of Eq.~(\ref{eq:dynamics_c}) is characterized by the
eigenvalues and eigenvectors of $\tilde{\vec{A}}$:
\begin{align}
\tilde{\vec{A}}\vec{Y}=\vec{Y}\bm{\Lambda}.
\label{eq:eigen_c}
\end{align}

With Eq.~(\ref{eq:dynamics_c}),
the encoded observation vector at timestep $t$
is given:
\begin{align}
  \bm{\psi}_{t}\approx\vec{A}^{t-1}\bm{\psi}_{1}
  +\sum_{\tau=1}^{t-1}\vec{A}^{t-\tau-1}\vec{B}\bm{\xi}_{\tau}.
\end{align}
Since $\vec{A}$ is approximated by $\vec{A}\approx\bm{\Phi}\bm{\Lambda}\bm{\Phi}^{\dagger}$
using the eigen decomposition result,
we can forecast the encoded observation vector at timestep $t$:
\begin{align}
  \hat{\bm{\psi}}_{t}=\bm{\Phi}\bm{\Lambda}^{t-1}\bm{\alpha}
  +\sum_{\tau=1}^{t-1}\bm{\Phi}\bm{\Lambda}^{t-\tau-1}\bm{\Phi}^{\dagger}\hat{\vec{B}}\bm{\xi}_{\tau}
  \label{eq:psihat_c}
\end{align}
where
\begin{align}
  \bm{\Phi}=\bm{\Psi}_{2}\vec{V}\bm{\Sigma}^{-1}\vec{U}_{1}^{\top}\hat{\vec{U}}\vec{Y},
\end{align}
and $\bm{\alpha}=\bm{\Phi}^{\dagger}\bm{\psi}_{1}$.
The observation vector at timestep $t$
is forecasted by $\hat{\vec{x}}_{t}=g(\hat{\bm{\psi}}_{t})$ by decoder $g$
in the same way as with Eq.~(\ref{eq:decoder}).
We estimate the parameters of the encoder and decoder neural networks
by minimizing the expected error in Eq.~(\ref{eq:Thetahat}).
Algorithm~\ref{alg:train_c} shows the estimation procedure
using a stochastic gradient descent-based method.

\begin{algorithm}[t!]
  \caption{Training procedures of the proposed method with control.}
  \label{alg:train_c}
  \begin{algorithmic}[1]
    \renewcommand{\algorithmicrequire}{\textbf{Input:}}
    \renewcommand{\algorithmicensure}{\textbf{Output:}}
    \REQUIRE{Observation time-series data $\vec{x}_{1},\cdots,\vec{x}_{T}$, exogenous time-series data $\vec{z}_{1},\cdots,\vec{z}_{T}$, batchsize $S$}
    \ENSURE{Estimated neural network parameters $\hat{\bm{\Theta}}$}
    \STATE Initialize parameters $\bm{\Theta}$.
    \WHILE{End condition is satisfied}
    \STATE Randomly sample timestep $\tau\in\{1,\cdots,T-S\}$.
    \STATE Encode observation vectors $\bm{\psi}_{t}=f(\vec{x}_{t})$ and
    exogenous vectors $\bm{\xi}_{t}=c(\vec{z}_{t})$ for $t\in\vec{t}=\{\tau,\cdots,\tau+S\}$.    
    \STATE Construct two encoded matrices $\bm{\Omega}=[[\bm{\psi}_{\tau};\bm{\xi}_{\tau}],\cdots,[\bm{\psi}_{\tau+S-1};\bm{\xi}_{\tau+S-1}]]\in\mathbb{R}^{(K+N)\times S}$ and $\bm{\Psi}_{2}=[\bm{\psi}_{\tau+1},\cdots,\bm{\psi}_{\tau+S}]\in\mathbb{R}^{K\times S}$ by Eqs.~(\ref{eq:Psi_c},\ref{eq:Omega})
    that start from $\tau$.
    \STATE Take the SVD of $\bm{\Omega}$ and $\bm{\Psi}_{2}$ in Eqs.~(\ref{eq:svd_Omega},\ref{eq:svd_psi2}).
    \STATE Calculate $\tilde{\vec{A}}$ by Eq.~(\ref{eq:Atilde_c}).
    \STATE Take the eigen decomposition of $\tilde{\vec{A}}$ in Eq.~(\ref{eq:eigen_c}).
    \STATE Forecast encoded observation vectors $\hat{\bm{\psi}}_{t}$ by Eq.~(\ref{eq:psihat_c}) for $t\in\vec{t}$.
    \STATE Forecast observation vectors $\hat{\vec{x}}_{t}$ by Eq.~(\ref{eq:decoder})  for $t\in\vec{t}$.
        \STATE Calculate loss
        $L=\frac{1}{S+1}\sum_{s=\tau}^{\tau+S}\parallel \bm{x}_{s}-\hat{\bm{x}}_{s}\parallel^{2}$, and its gradient.
    \STATE Update parameters $\bm{\Theta}$ using the loss and the gradient.
    \ENDWHILE
  \end{algorithmic}
\end{algorithm}

\section{Experiments}
\label{sec:experiments}

\subsection{Eigenvalue estimation}

\begin{figure*}[t!]
\centering   
{\tabcolsep=-1.0em
\begin{tabular}{ccc}
 \multicolumn{3}{c}{Observation time-series data} \\
  \includegraphics[width=16em]{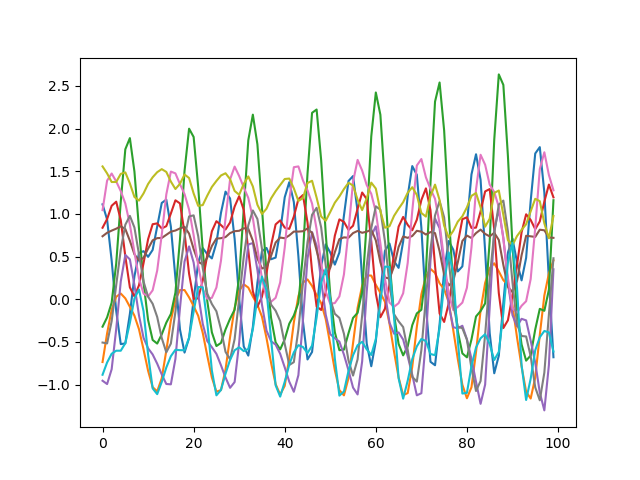}&
  \includegraphics[width=16em]{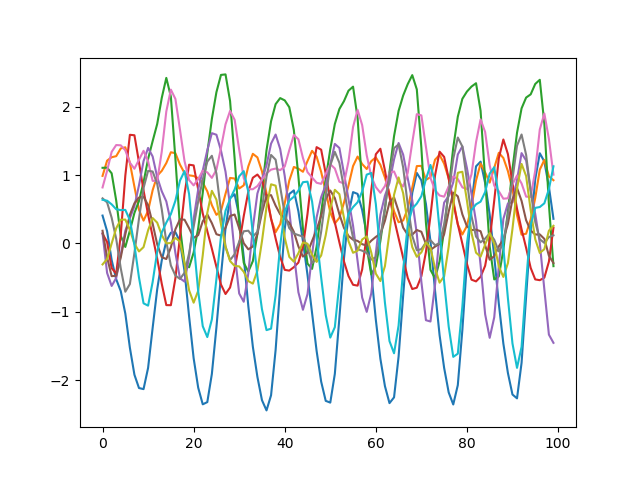}&
  \includegraphics[width=16em]{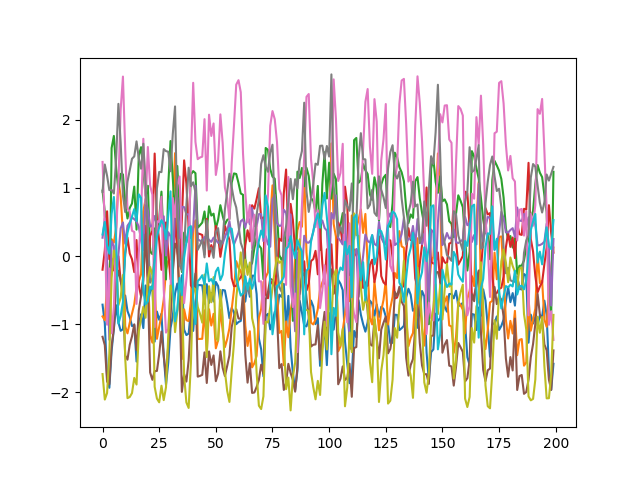}\\
 \multicolumn{3}{c}{True and estimated eigenvalues}\\
  \includegraphics[width=18em]{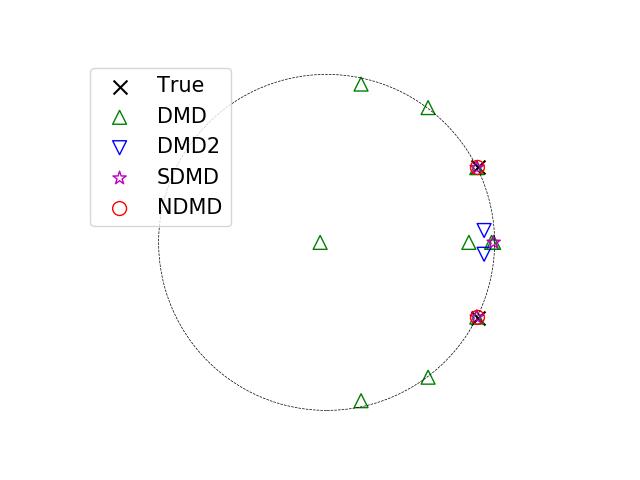}&  
  \includegraphics[width=18em]{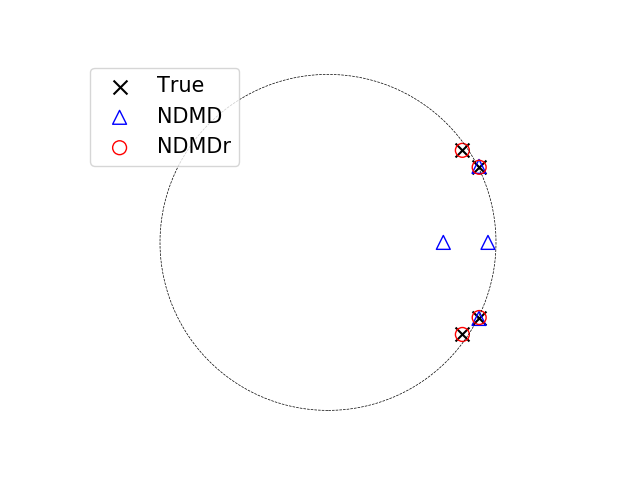}&
  \includegraphics[width=18em]{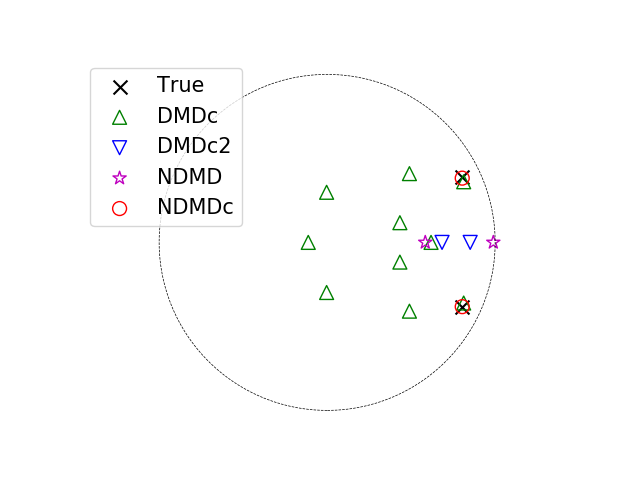}\\
    {\small (a) Observation time-series} & {\small (b) w/ auxiliary information} & {\small (c) w/ exogenous time-series}\\
\end{tabular}}   
\caption{Observation time-series data, and true and estimated eigenvalues for eigenvalue estimation experiments with (a) observation time-series data, (b)  observation time-series data and auxiliary information, and (c) observation and exogenous time-series data.}
  \label{fig:synth}
\end{figure*}

We evaluated the proposed NDMD described in Section~\ref{sec:ndmd} using synthetic data,
which were generated by the following procedure.
First, the dynamics in a two-dimensional lifted space was obtained
with linear transition matrix,
$\vec{A}=[0.9,-0.5; 0.4,0.9]$.
Then the dynamics in a ten-dimensional observation space was calculated
by a nonlinear transformation using Gaussian processes (GPs)~\cite{rasmussen2005gaussian}
with the RBF kernel.
The time-series in the observation space are shown in Figure~\ref{fig:synth}(a, top).
The first 70 timesteps of the data were used for training,
and the following 10 were used for validation.

We compared the eigenvalue estimation performance of the proposed method (NDMD)
with DMD~\cite{tu2013dynamic}, a DMD with rank-two approximation (DMD2),
and a sparse DMD~\cite{jovanovic2014sparsity} (SDMD).
With the proposed method, we used
four-layered feed-forward neural networks
with 256 hidden units for the encoder and decoder.
The dimensionality of the lifted space was two.
We optimized the parameters using Adam~\cite{kingma2014adam} with a learning rate of $10^{-3}$,
a dropout rate of $0.1$~\cite{srivastava2014dropout}, and a batch size of 128.
We implemented the proposed method with Tensorflow~\cite{tensorflow2015-whitepaper}.
The validation data were used for early stopping,
and the maximum number of epochs was 1,000.
With DMD2, we selected two eigenvalues based on the singular values.
With SDMD, we tuned the best hyperparameter
from $\{10^{-3},10^{-2},10^{-1},1,10^{1},10^{2},10^{3}\}$.

Figure~\ref{fig:synth}(a, bottom) shows the true and estimated eigenvalues.
The proposed NDMD appropriately estimated the two eigenvalues. 
The DMD obtained ten eigenvalues. Although two were close to the true eigenvalues,
the other eight were different from the true eigenvalues.
The DMD2 estimations were far from the true values.
The SDMD obtained three eigenvalues, only two of which were close to the true eigenvalues.
This result indicates that the proposed NDMD
can obtain sparse dynamics properly using neural networks.

\subsection{Eigenvalue estimation with auxiliary information}

We evaluated the proposed NDMD with auxiliary information described in Section~\ref{sec:ndmda} using synthetic data.
The dynamics in a four-dimensional lifted space was obtained
with linear transition matrix $\vec{A}=[0.9,-0.5,0,0;0.4,0.9,0,0;0,0,0.8,-0.5;0,0,0.6,0.8]$,
and the dynamics in a ten-dimensional observation space was calculated
using GPs with an RBF kernel.
The observation time-series data are shown in Figure~\ref{fig:synth}(b, top).
The first 70 timesteps of the data were used for training,
the next 10 were used for validation,
and the remaining 20 were used for testing.

We calculated the four true eigenvalues from $\vec{A}$, and used them for auxiliary information.
Figure~\ref{fig:synth}(b, bottom) shows the result.
The NDMD without auxiliary information (NDMD)
failed to estimate two eigenvalues in this case.
On the other hand, the NDMD with auxiliary information (NDMDr)
appropriately estimated all four eigenvalues.
The mean squared error for forecasting the test data
was 0.146 with NDMD and 0.074 with NDMDr.
This result indicates that the proposed method can improve
the forecasting performance using auxiliary information.

\subsection{Eigenvalue estimation with control}

We evaluated the proposed NDMD with control described in Section~\ref{sec:ndmdc} using synthetic data.
The data were generated by the following procedure.
First, a one-dimensional exogenous time-series was obtained
by a standard Gaussian distribution.
Second, the dynamics in a two-dimensional lifted space was obtained
with linear transition matrices $\vec{A}=[0.9,-0.5;0.4,0.9]$,
and $\vec{B}=[1;0]$.
Finally, we calculated the dynamics in a ten-dimensional observation space 
by nonlinear transformation using GPs with the RBF kernel.
The observation time-series data are shown in Figure~\ref{fig:synth}(c, top).
The first 140 timesteps of the data were used for training,
and the next 20 were used for validation.

Figure~\ref{fig:synth}(c,bottom) shows the estimated eigenvalues.
The neural DMD with control (NDMDc) successfully estimated the true eigenvalues.
The NDMD without control (NDMD) failed to estimate them
since it ignores the influence of the exogenous time-series data.
The DMD with control (DMDc) estimated ten eigenvalues,
only two of thewhich were close to the true ones.
The estimated eigenvalues by the DMD with control with rank-two approximation (DMDc2)
were different from the true ones.

\subsection{Fluid flow forecast}

\begin{figure}[t!]
  \centering      
  \includegraphics[width=18em]{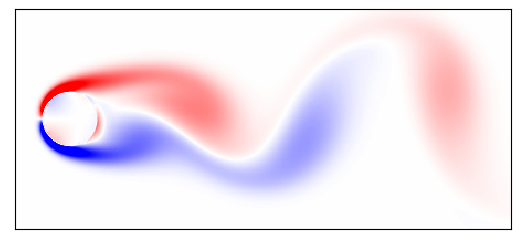}
  \caption{Cylinder wake data for fluid flow forecast.}
  \label{fig:cylinder}
\end{figure}  

\begin{table*}[t!]
\centering
\caption{Average mean squared error and its standard error for fluid flow forecast.}
\label{tab:cylinder}
{\tabcolsep=0.3em\begin{tabular}{lrrr}
\hline
& 0.01\% & 0.03\% & 0.1\%\\
\hline
NDMD & {\bf 0.000211$\pm$0.000097}  & {\bf 0.000169$\pm$0.000129}  & {\bf 0.000029$\pm$0.000004}  \\
DMD &  0.332469$\pm$0.037414  &  0.011987$\pm$0.004589  &  0.000292$\pm$0.000052  \\
EDMD &  0.080878$\pm$0.032596  &  0.001899$\pm$0.001318  &  0.000055$\pm$0.000016  \\
KDMD(RBF) &  0.027164$\pm$0.011372  &  0.000435$\pm$0.000237  &  0.000173$\pm$0.000063  \\
KDMD(Poly) & 0.311673$\pm$0.054852  &  0.039735$\pm$0.015964  &  0.002097$\pm$0.000647  \\
AEAR & 0.027269$\pm$0.016137  &  0.001555$\pm$0.001245  &  0.000100$\pm$0.000028  \\
LKIS &  0.023177$\pm$0.013134  &  0.001525$\pm$0.001257  &  0.000101$\pm$0.000027  \\
AR &  59204.80$\pm$49533.98  &  20.59142$\pm$17.15388  &  0.109702$\pm$0.087929  \\
NN &  0.011549$\pm$0.006359  &  0.005865$\pm$0.005476  &  0.000036$\pm$0.000010  \\
LSTM &  0.157520$\pm$0.045537 &  0.012063$\pm$0.004472 &  0.000320$\pm$0.000042\\
\hline
\end{tabular}}
\end{table*}

We evaluated our proposed NDMD with fluid flow forecasts.
We used the cylinder wake data shown in Figure~\ref{fig:cylinder}.
The original data consisted of the vorticity in $199\times499$ field for 151 timesteps.
We randomly subsampled $\{0.01\%,0.03\%,0.1\%\}$ of the field for the observation time-series data,
and forecasted them.
We used 70\% of the timesteps for training, 10\% for validation, and the remainder for testing.

We compared NDMD with the following nine methods:
DMD, EDMD, KDMD(RBF), KDMD(Poly), AEAR, LKIS, AR, NN, and LSTM.
With NDMD, we used four-layered neural networks with 256 hidden units and the 256 dimensional lifted space,
for the encoder and decoder.
The rank of the linear transition in the lifted space was determined
by removing the elements that had relative singular values lower than 0.1\%.
DMD is the standard DMD without neural networks.
EDMD is the extended DMD~\cite{williams2015data}, where observation vectors were augmented by
the product of two elements in the original observation vector.
KDMD is the kernel DMD~\cite{williams2015kernel}, where we used the RBF kernel
and the polynomial kernel with a degree of three.
AEAR is a combination of neural network-based autoencoders and linear autoregressive models~\cite{lusch2018deep}.
It forecasted the value at the next timestep
using an encoder, a linear transition matrix, and a decoder.
The observation vectors were first encoded by the encoder.
Next, the encoded vectors were transformed to those at the next timestep
by the linear transition matrix.
Then, the transformed encoded vectors were decoded in the observation space.
The neural networks were trained by minimizing the mean squared error for the next timestep forecast.
LKIS is learning Koopman invariant subspaces~\cite{takeishi2017learning},
which also consisted of an encoder, a linear transition matrix, and a decoder.
LKIS was trained by minimizing the sum of the reconstruction error in the observation space,
and the next timestep forecasting error in the lifted space.
AR is the linear autoregressive model.
NN is a neural network that forecasts the next timestep observation vectors.
We used four-layered neural networks with 256 hidden units.
LSTM is a long short-term memory recurrent neural network~\cite{hochreiter1997long},
where 256 hidden units were used.
AR, NN, and LSTM were trained by minimizing the mean squared error for the next timestep forecast.

The test mean squared error averaged over ten experiments with different random subsamplings is
shown in Table~\ref{tab:cylinder}.
The proposed NDMD achieved the lowest error.
As the number of observation fields increased,
the error decreased.
The DMD failed to produce a forecast since it could not extract the nonlinear dynamics.
Although EDMD and KDMD were outperformed DMD by incorporating higher order information,
they were worse than NDMD
since the predefined nonlinear functions were not appropriate for the dynamics.
On the other hand, the proposed method learned a proper lifted space from the data by neural networks.
The errors by existing neural network-based methods, i.e., AEAR, LKIS, NN, and LSTM, were higher than NDMD.
The reason is that NDMD was trained based on long-term forecasting performance using the eigenvalues,
and the noise was removed by SVD.
The forecasting values by AR diverged,
and the error was very high, especially when there were few observations.

\section{Conclusion}
\label{sec:conclusion}

We proposed a neural network-based method for analyzing nonlinear dynamics by
the Koopman operator.
With the proposed method, we estimated
a lifted space with a linear dynamics
such that
the forecast error is minimized
when the dynamics is modeled based on spectral decomposition in the lifted space.
We demonstrated the effectiveness of the proposed method
for estimating the eigenvalues of the nonlinear dynamics and forecasting future values
compared with existing DMD-based and neural network-based methods.

Although our results are encouraging, we must 
extend our approach in several directions,
First, we will apply it for controlling the dynamics. 
Since the proposed method can estimate lifted linear dynamics,
we can control the nonlinear dynamics using techniques developed in optimum control theory for linear dynamics~\cite{kirk2004optimal,hao2020data,morton2018deep,li2020learning},
such as linear quadratic control~\cite{ji1990controllability}
and pole placement~\cite{chilali1999robust}.
Second, we want to use other types of neural networks, which include
convolutional neural networks~\cite{krizhevsky2012imagenet}
and transformers~\cite{vaswani2017attention},
for extracting rich information from data.
Third, we will use our proposed end-to-end learning approach for
supervised learning tasks, such as time-series regression and classification~\cite{fujii2019dynamic}.
Finally, we plan to evaluate our method with various types of auxiliary information,
e.g., a regularizer on dynamic modes.

\bibliographystyle{abbrv}
\bibliography{siam2020}

\end{document}